\pdfoutput=1

\documentclass[11pt]{article}

\usepackage[]{acl}

\usepackage{times}
\usepackage{latexsym}

\usepackage{hyperref}
\usepackage{url}
\usepackage{multirow}
\usepackage{graphicx}
\usepackage{comment}
\usepackage{amssymb}
\usepackage{amsmath}
\usepackage{url}
\usepackage{subfigure}

\usepackage[T1]{fontenc}

\usepackage[utf8]{inputenc}

\usepackage{microtype}

\usepackage{inconsolata}

\usepackage{graphicx}

\usepackage{enumerate}
\usepackage{amsmath}
\usepackage{microtype}
\usepackage{hyperref}
\usepackage{url}
\usepackage{booktabs}
\usepackage{listings}
\title{CTP-LLM: Clinical Trial Phase Transition Prediction Using Large Language Models}

  \author{
Michael Reinisch{$^{1}$}, Jianfeng He{$^4$}\thanks{~~The work was done prior to joining Amazon.}~, Chenxi Liao{$^2$}, Sauleh Ahmad Siddiqui{$^3$}, Bei Xiao{$^1$}
\\ 
   {$^1$} Department of Computer Science, American University, Washington, DC, USA\\
    {$^2$} Department of Neuroscience, American University, Washington, DC, USA\\
    {$^3$} Department of Environmental Science, American University, Washington, DC, USA\\
 {$^4$} Department of Computer Science, Virginia Tech, Falls Church, VA, USA\\
$^1${bxiao@american.edu}
 }

\begin{document}
\maketitle

\begin{abstract}
New medical treatment development requires multiple phases of clinical trials. Despite the significant human and financial costs of bringing a drug to market, less than 20\% of drugs in testing will make it from the first phase to final approval. Recent literature indicates that the design of the trial protocols significantly contributes to trial performance. We investigated Clinical Trial Outcome Prediction (CTOP) using trial design documents to predict phase transitions automatically. We propose CTP-LLM, the first Large Language Model (LLM) based model for CTOP. We also introduce the PhaseTransition (PT) Dataset; which labels trials based on their progression through the regulatory process and serves as a benchmark for CTOP evaluation. Our fine-tuned GPT-3.5-based model (CTP-LLM) predicts clinical trial phase transition by analyzing the trial's original protocol texts without requiring human-selected features. CTP-LLM achieves a 67\% accuracy rate in predicting trial phase transitions across all phases and a 75\% accuracy rate specifically in predicting the transition from Phase~III to final approval. Our experimental performance highlights the potential of LLM-powered applications in forecasting clinical trial outcomes and assessing trial design.

\end{abstract}

\section{Introduction}
\label{sec:intro}


\begin{figure}[ht]
\begin{center}
\includegraphics[width=2.2in]{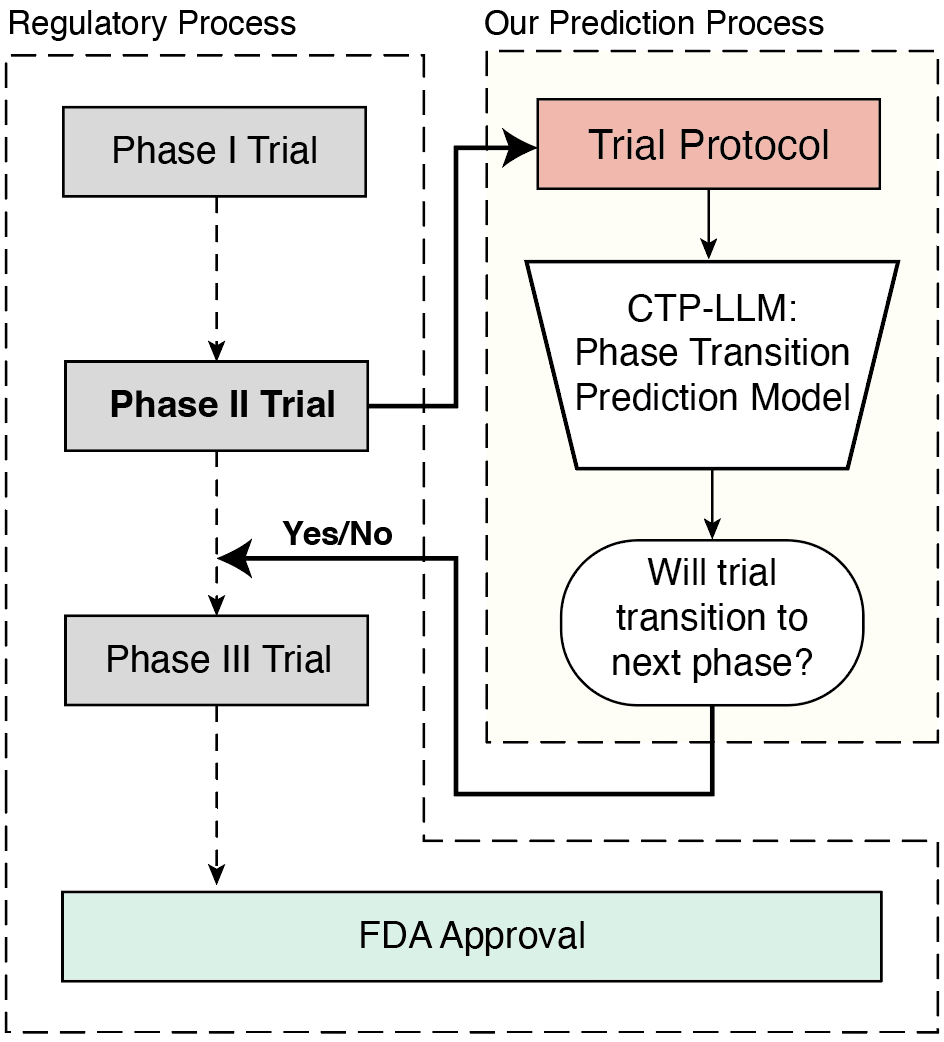}
\end{center}
\caption{Overview of our framework, CTP-LLM, for clinical trial phase transition prediction. A trial protocol is a comprehensive document that outlines the plan for conducting the trial (see Appendix \ref{sec:app_protocol}). A treatment is typically tested in three phases, starting with safety evaluation and dosage in Phase I with a small group of people, then assessing efficacy in Phase II with a larger group, and finally confirming efficacy and safety in Phase III with a large population before FDA approval. However, the treatment can drop out in any phase. Our model takes a protocol of a given phase as input and predicts whether it can successfully transition to the next phase before the trial starts. \label{fig:title}}
\end{figure}
A clinical trial systematically evaluates the safety and efficacy of medical interventions on human subjects, categorized into Phases I, II, and III, and the final FDA approval. High drug attrition in the regulatory process is well-documented in the literature~\cite{hay2014clinical,kola2004can, pammolli2011productivity}. For example, 60–70\% of Phase II trials and 30–40\% of Phase III trials failed for the next phase~\cite{feijoo2020key}. This leads to extraordinary costs for bringing a drug to market, with estimates ranging from \$600 million to \$2.8 billion~\cite{feijoo2020key, sertkaya2016key}. This paper focuses on predicting Clinical Trial Phase Transition from Trial Protocols, an early forecasts for clinical trial success, which assists trial designers make more informed decisions about the design and allocate resources efficiently.



All clinical trial phases follow a textual protocol outlining the study objectives, treatment plan, participant recruitment criteria, and procedures to be followed during the study (see~\ref{sec:app_protocol} for the detailed description of the trial protocol). Although the primary objective of a trial is to assess a drug's efficacy, effectiveness, and safety, it is widely recognized that the protocol design is crucial to the success of clinical trials~\citep{chow2008design, getz2008assessing, vischer2017increasing}.  Qualitative evidence from stakeholders has shown that protocol design is critical for a drug's successful progression through the phases of the clinical trials~\citep{chow2008design, getz2008assessing, vischer2017increasing}. Research suggests that trial protocol complexity, the type of company performing the trial, and barriers erected by regulatory agencies contribute to trial failure~\citep{dimasi2015tool, getz2008assessing, getz2014, hay2014clinical, kola2004can, pammolli2011productivity, ringel2013does, scannell2012diagnosing}. In order to automatically discover factors contributing to clinical trial outcomes in timely manner, our paper proposes two large language models to predict clinical trials from one phase to the next based on protocol-time known variables (see Figure \ref{fig:title}), an under-explored area~\citep{feijoo2020key}. 


There is currently no effective model that automatically forecast clinical trial phase transition by taking the protocol design documents as input. This is mainly due to two challenges. The first challenge is the dataset used lacks correct phase transition labeling. Previous work used the trial recruitment status (e.g., ``Completed" or ``Terminated") as a measure of success 
~\cite{ferdowsi2023deep, follett2019quantifying, fu2022hint,luo2023clinicalrisk}. It only indicates if the trial was completed as scheduled but do not predict if it successfully transitioned to the next phase. This renders the predictions of existing models inaccurate. The second issue of previous work is their dependence on human-selected features. Clinical Trial failure is a complex topic, dependent on multiple variables \cite{feijoo2020key, ferdowsi2023deep, friedman2015fundamentals}. A human expert would have difficulty determining which variables influence the phase transition due to the need for in-depth knowledge across all medical fields. Therefore, a data-driven approach automatically identifying relevant features from trial documents is advantageous for creating a successful prediction and can be better evaluated later by experts. 

To tackle these challenges, we propose two models for predicting trial phase transition success or failure using multiple documents as input, eliminating the need for human-selected features and inference (see Figure \ref{fig:labelling} and section \ref{sec:labelling} for details). Our first model, CTP-LLM, is a specialized version of GPT-3.5 Turbo, trained on our PhaseTransition (PT) Dataset. Our second model, BERT+RF, combines a clinical Bidirectional Encoder Representations from Transformers (BERT) with a Random Forest (RF). BERT+RF offers the advantage of low computational cost and can be trained on most machines, while CTP-LLM achieves higher performances overall. Our proposed models show robust accuracy in predicting phase transitions. The CTP-LLM model achieves a F1 score of 0.67 when integrating trial information across all phases. When specifically trained to predict transitions to Phase~III \text{-} the most costly phase in the regulatory process \text{-} the model demonstrates even higher accuracy with an F1 score of 0.75.

Our contributions are as follows: 
\begin{itemize}
    \item \textbf{Establish the framework of using language models including fine-tuned LLM in Automatic CTOP.} We are the first to leverage the capabilities of LLMs for CTOP and introduce a benchmark for predicting phase transitions of clinical trials as an indication of trial outcome.

    \item \textbf{Build a new open-source dataset for CTOP}. We introduce the PT Dataset, a new resource specifically designed for clinical trial phase transition prediction. The dataset links trial protocols with information on whether the trial advanced to the next phase. 

    \item \textbf{A new benchmark for CTOP evaluation.} We present an improved task for CTOP and experimentally demonstrate that our proposed methods accurately predict clinical trial phase transitions. We also discovered features critical for failures of phase transition. Thus, we provide a solid foundation for further advancements in CTOP. 

\end{itemize}

\section{Related Work} 
\label{sec:related-work}
Data mining on Clinical trials is an emerging research topic with a small set of relevant literature. Some previous work aimed to reduce the attrition rate in clinical trials by classifying eligibility criteria to select more suitable participants \cite{li2022comparative, tian2021transformer, zeng2021automated}). However, this approach disregards flaws in the trial protocol, while our approach aims to identify disadvantageous trial protocols.\par
Other approaches focus on CTOP. \cite{artemov2016integrated} and \cite{gayvert2016data} link the outcome of the trial to drug toxicity and side effects, \cite{jin2020predicting, wang2022artificial, wang-sun-2022-trial2vec}) try to improve trial design through simulation, \cite{follett2019quantifying} quantify the risk of trial termination through text mining, while \cite{qi2019predicting} leverage deep learning for CTOP by analyzing pharmacokinetic concentrations and connecting them to patient characteristics. However, the effectiveness of these existing CTOP methods is affected by similar limitations, such as training on data that only becomes available during or after the trial \cite{feijoo2020key, ferdowsi2023deep}, relying on human-selected features that do not generalize well to textual data \cite{feijoo2020key, fu2022hint, kavalci2023improving}, focusing only on predictions for specific diseases and phases \cite{aliper2023prediction, feijoo2020key}, or being only applicable to molecular drugs with a publicly available chemical structure \cite{fu2022hint, qi2019predicting}. Differently, our approach is applicable to trials of all phases and treatment types.\par
Moreover, the majority of the aforementioned approaches share a commonality in labeling trials based on their completion or termination status, which, in reality, does not serve as a reliable indicator of trial success. Ferdowsi et al. proposed a risk assignment methodology based on historical clinical trial statistics as well as termination status to label the trial protocols \cite{ferdowsi2023deep}. In contrast, we label the trials based on trial phase transitions with an additional dataset that tracks the phase progression of the trials. \par

So far, only \cite{feijoo2020key} have attempted to use phase transitions to indicate success for CTOP. However, their model relies on human-selected features, which does not allow the protocol documents as input. Our approach predicts phase transitions directly from the trial protocol documents.\par
The utilization of LLMs in the clinical trial domain has been sparse, with only a few published papers focusing on different aspects of the domain, such as matching patients to trials or data summarization \cite{jin2023matching, wang2023autotrial, white2023clinidigest, zheng2024multimodal}. We are the first to present an approach that uses an LLM for CTOP.

\section{Method}
\begin{figure*}[!ht]
\begin{center}
\includegraphics[width=6in]{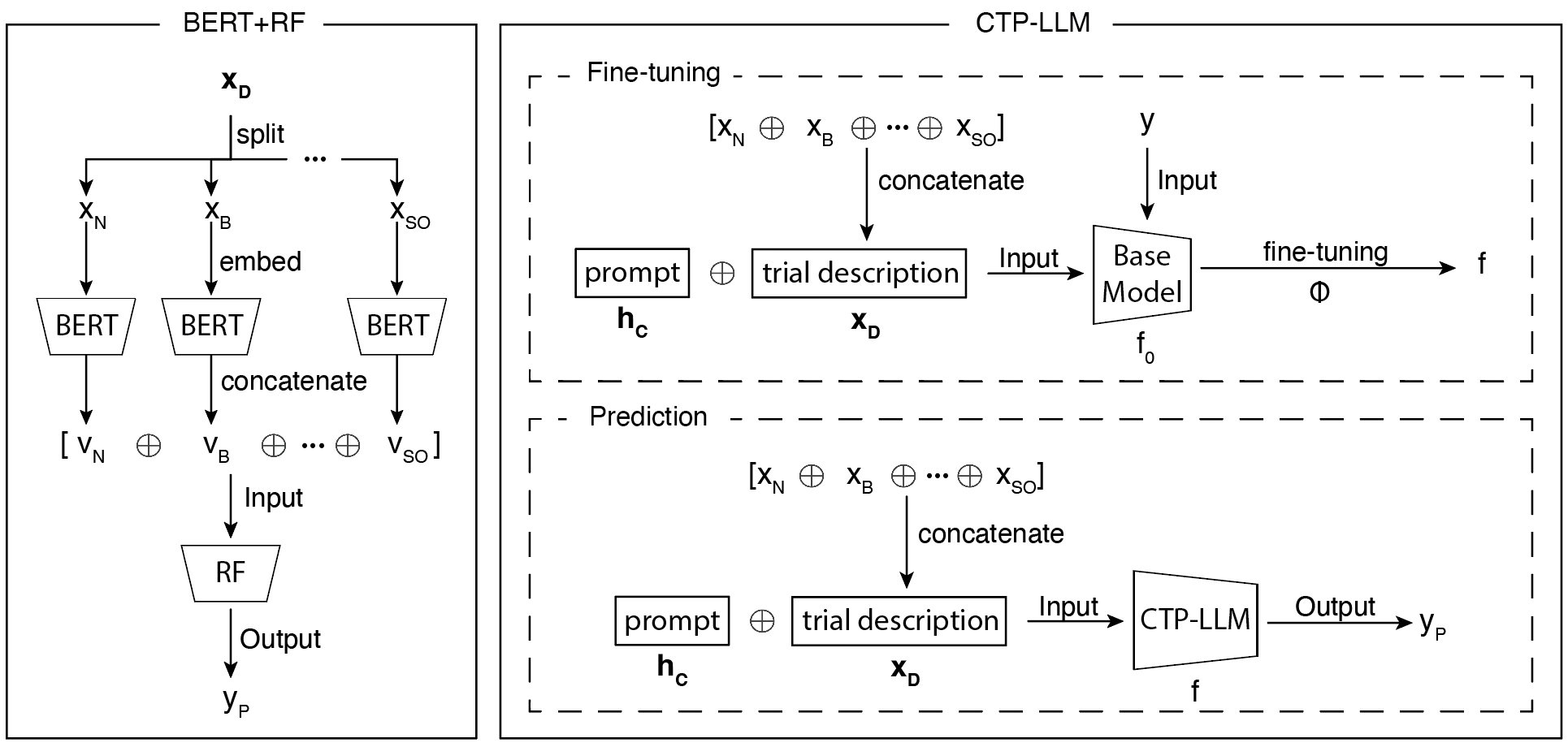}
\end{center}
\caption{Overview of the two models. On the left is the BERT+RF approach, where the trial textual description $x_D$ is divided into its entries, individually embedded by the clinical BERT, concatenated, and then inputted into the RF classifier. On the right are the two steps of the CTP-LLM approach. First, the instruction fine-tuning of the base model $f$, using trial description $x_D$, the prompt $h_C$, and the labels $y$ as inputs to the fine-tuning function $\Phi$, resulting in the fine-tuned model, CTP-LLM ($f$). For an example of the prompt, refer to Table \ref{tab:prompt} in Appendix. CTP-LLM only requires the prompt and a trial description to generate a prediction. \label{fig:models}}
\end{figure*}

Figure~\ref{fig:models} illustrates our methods for constructing two language models for clinical trial phase transition prediction. It contains two independent models: the BERT+RF construction and the CTP-LLM fine-tuning and prediction process.

\subsection{Problem Setup}
Our approach aims to develop a model, represented by a function $f\in\mathbb{Z}_{2}$, that predicts the target trial outcome $y_p\in\mathbb{Z}_{2}$ based on the input $x_D$. Here, $x_D$ denotes the trial description, a concatenation of several data elements. The approach consists of two stages: \textbf{dataset creation} and \textbf{model training}. 

\begin{itemize}
    \item Dataset creation refers to compiling an accurate CTOP dataset, providing $x_D$, that includes trial protocol data labeled with phase transition information inferred from drug performance data. Every trial that enters each phase has a unique protocol.
    \item In model training, both our models (BERT+RF and CTP-LLM) are trained to predict the target phase transition according to a given input trial description, as described above. 
\end{itemize}

It is essential to highlight that the trial description is solely derived from textual data obtained from the publicly accessible \textit{ClinicalTrials.gov} database. All used texts are created in the trial design process before the trial starts, ensuring that our models can effectively predict the outcome of a trial before approval from the Food and Drug Administration (FDA) is sought.

\subsection{Model Overview} \label{sec:model-overview}
We introduce two models for the phase transition prediction task that differ in architecture, performance, and complexity. We have chosen an LLM approach, as it represents the state-of-the-art in NLP, and a combined deep and shallow learning approach, chosen for its accessibility and suitability for training on a wide range of computing setups. These models are:

\paragraph{BERT+RF} 
A hybrid method that enhances predictive accuracy by aggregating predictions from multiple decision trees. It excels through its lower computational requirements compared to the LLM. One limitation for our specific use case is BERT models' restricted attention window size, typically allowing text up to 512 tokens \citep{devlin2018bert, dai2022revisiting}. To overcome this limitation, we employ a hybrid approach by combining clinical BERT embedding with Random Forest (RF) classification. Each trial protocol's different attributes (e.g., name, description, recruitment criteria) are embedded separately using a clinical BERT model \citep{rohanian2023lightweight}, resulting in numerical representations. These representations are then concatenated into a high-dimensional feature vector on which we train an RF classifier (see Section \ref{sec:bert-rf}).
    
\paragraph{CTP-LLM} As stated in Section \ref{sec:intro} a successful prediction model has to detect relevant aspects across trial protocols automatically. LLMs excel in discovering patterns and nuances across textual data and are preferable for our task \citep{thirunavukarasu2023large, ye2023comprehensive}. While dedicated medical large language models are available, \citet{zhou2024survey} demonstrated that GPT-3.5 Turbo exhibits robust performance on medical downstream tasks. The model was fine-tuned by concatenating 11 high-quality attributes extracted from the trial protocol, accompanied by a prompt. The prompt aims to instruct the model on the tasks and ensures its output is binary (refer to Table \ref{tab:prompt}).


\subsection{PhaseTransition Dataset Construction} \label{sec:dataset-creation}
To construct a dataset that labels trials according to their advance through the regulatory process, two aspects are essential: \textbf{gathering clinical trial protocols} and \textbf{establishing connections between medical treatments across trials}. Two distinct sources are used to gather the relevant information:

\begin{itemize}
    \item \textbf{ClinicalTrials.gov} (\href{https://clinicaltrials.gov/}{http://clinicaltrials.gov/}) is an English-language clinical repository in the public domain maintained by the United States National Library of Medicine, offering comprehensive data on clinical studies worldwide. Presently, it houses 481,198 study records from 223 countries, making it the largest database of its kind globally.
    \item \textbf{Biomedtracker} (\href{https://www.biomedtracker.com/}{https://www.biomedtracker.com/}) is an English-language proprietary database compiled by Informa Business Intelligence Inc. It is a comprehensive resource that tracks and analyzes pharmaceutical and biotechnology industry developments. 
    Biomedtracker allows us to track a treatment's performance through multiple clinical studies. The version we used contains information on 20,016 unique drugs.
\end{itemize}

By merging trial information from Biomedtracker and ClinicalTrials.gov based on the common National Clinical Trial Identifier (NCT-ID) and excluding low-quality trials, we obtained an initial dataset comprising 20,000 entries.

\begin{figure*}[!ht]
\begin{center}
\includegraphics[width=5.3in]{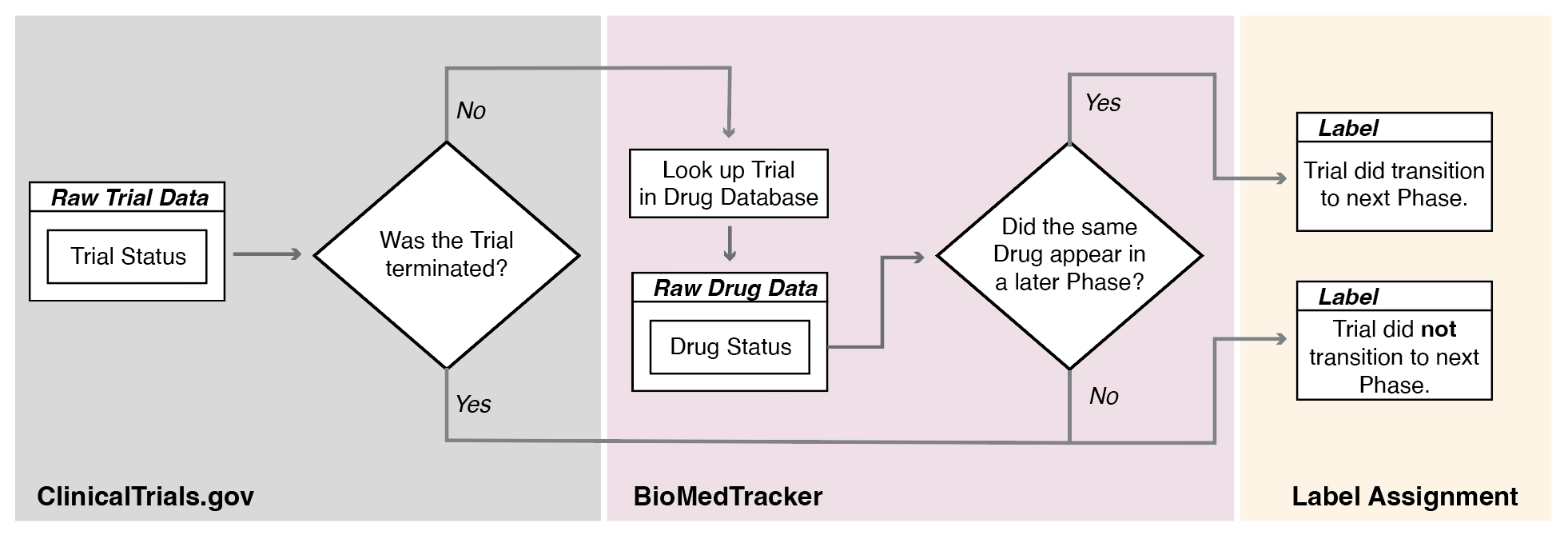}
\caption{Overview of the labelling process as described in Section \ref{sec:labelling}. \label{fig:labelling}}
\end{center}
\end{figure*}

\textbf {Labelling process} \label{sec:labelling}
The common indicator of trial success or failure in previous CTOP work is the trial (recruitment) status. It can have several values, whereby it is most often broken down to `completed' or `terminated.' However, none of these labels provides information on a trial's transition to the next phase. A trial can have been completed as planned while having proven that the test treatment has no efficacy \citep{feijoo2020key}. Similarly, a trial could have been terminated because the tested treatment showed promising effects earlier than expected \citep{feijoo2020key}. Therefore, the recruitment status alone is insufficient to label a trial accurately. \par

While the same medical intervention can be tested for multiple medical conditions, we only consider trials in different phases to be related if they feature the same intervention and target the same disease. This information is captured by the Drug-Indication ID. By using the Biomedtracker, we can successfully connect individual trials based on this indicator, which enables labeling each one of them based on three distinct rules:

\begin{enumerate}
    \item \textbf{Successful Phase Transition}: If, according to Biomedtracker, a drug advances to a certain ultimate phase, all trials in preceding phases featuring the same Drug-Indication ID have performed a successful phase transition.

    \item \textbf{Incomplete Phase Information}: If, according to Biomedtracker, a drug advances to a certain ultimate phase, all trials of this phase featuring the same Drug-Indication ID have not performed a successful phase transition.

    \item \textbf{Unsuccessful Phase Transition}: Trials labeled as terminated on ClinicalTrials.gov have not performed a successful phase transition, even if indicated otherwise by the Biomedtracker.

\end{enumerate}

See Figure \ref{fig:labelling}. While the third rule contradicts our earlier assertion that terminated trials can still result in a successful phase transition, we have opted against labeling these trials as such. Since the model is not given information on which trials are connected, Rule 3 further ensures that trial labels are not sequential, preventing model bias when training on all phases at once.



\subsubsection{Data Synthesis} \label{sec:synthesis}
After assigning each trial an accurate phase transition label, the next step is preparing the trial protocols for processing. Even though we want to give the model the full trial protocol and let it automatically identify relevant attributes, some human intervention is still necessary. We filtered out data that was only available after the trial had started, such as the number of participants, to prevent any potential look-ahead bias. ClinicalTrial.gov entries are of varying quality \citep{tse2018avoid}. Therefore, we choose attributes (a subset of documents) of the protocols that are of high quality (e.g., present, complete, informative) throughout most trial entries, such as name, brief description, and recruitment criteria. For a list of all selected attributes, see Table \ref{tab:attributes} and \ref{sec:feature_description} for their explanation.


By concatenating these attributes, we create the trial description $x_D$ according to

\begin{equation}
    x_D = (x_N \oplus x_B ... \oplus x_{SO}). \label{eq:concat-data}
\end{equation}
\\
An example of a trial description can be seen in Table \ref{tab:data-point}. Let $\mathcal{D}$ be the resulting PT dataset, where each row consists of the input $x_D$, with its corresponding phase transition label $y$, either \textit{Yes}, for a successful phase transition, or \textit{No} for not having performed a phase transition. We can represent $\mathcal{D}$ as a set of ordered pairs as

\begin{equation}
    \mathcal{D} = \{(x_{D1}, y_{1}), (x_{D2}, y_{2}), ..., (x_{Dn}, y_{n})\}, \label{eq:dataset}
\end{equation}
\\
with each ordered pair $\mathcal{D}_i = (x_{Di}, y_{i})$ representing a data point. Table \ref{tab:data-point} shows an example of a data point.

\subsection{Model Training}
After giving a global view of our two models in Section \ref{sec:model-overview} and describing the creation of the PT Dataset in Section \ref{sec:dataset-creation}, in this section, we now detail how we trained each model.

\subsubsection{BERT+RF} \label{sec:bert-rf}
Let $x_N = [x_{N_1}, x_{N_2}, ..., x_{N_d}]$ be the name of the clinical trial, where $x_{N_i}$ represents the $i$-th token in the text, and $d$ the last token of the text. We compute the embedding of the trial name by using a clinical BERT model as

\begin{equation}
    v_N = BERT(x_N) .
\end{equation}
\\
With $v_N =  \in \mathbb{R}^h$, where $h=768$, due to the BERTs defined embedding size. The embedding process is repeated for each data element in $x_D$, with the resulting embedding vectors for each  document being concatenated similarly to Equation \ref{eq:concat-data} as

\begin{equation}
    v_D = (v_N \oplus v_B \oplus ... \oplus v_{SO}).
\end{equation}
\\
Thus, $v_D = [v_{D_1}, v_{D_2}, ..., v_{D_{11h}}]$ be the embedding feature vector of $x_D$. By representing the associated label $y_{B}$ as a binary numerical value  according to 

\begin{equation}
    y_{B} = \begin{cases} 
    1, & \text{if label } y = \text{"Yes"} \\
    0, & \text{if label } y = \text{"No"} 
    \end{cases} ,
\end{equation}
\\
where, $B$ could be any trial transition from I$\,\to\,$II,II$\,\to\,$III. We can rewrite the dataset used for the RF classifier as 

\begin{equation}
    \mathcal{D}_{RF} = \{(v_{D1}, y_{B1}), ..., (v_{Dn}, y_{Bn})\}. 
\end{equation}
\\
Following the RF algorithm, we randomly select $m$ data points with replacement from the dataset $\mathcal{D}_{RF}$ to create $B=100$ bootstrap samples $\mathcal{D}^*_b$, where $b=1,2,...,B$. For each bootstrap sample $\mathcal{D}^*_b$ a decision tree $T_b$ is grown from a random subset of features at each split until all leaves are pure, with Gini impurity being the splitting criterion. For a node $t$ in the decision tree, if $p_i$ represents the proportion of samples of class $i$ in node $t$, then the Gini impurity $G(t)$ is defined as
\begin{equation}
    G(t) = 1 - \sum_{i=1}^{C} p_{i}^{2},
\end{equation}
\\
where $C=2$ being the number of classes. Finally, the predictions of all decision trees are aggregated using the majority voting aggregation function $Agg$, and the predicted phase transition label $y_p$ is calculated by

\begin{equation}
    y_p = \text{Agg}(\{ T_1, T_2, ..., T_B \}, v_D).
\end{equation}

\subsubsection{CTP-LLM} \label{sec:chat-ctp}
In contrast to the BERT+RF model, which we train from scratch, CTP-LLM is created by instruction fine-tuning GPT-3.5 Turbo on $\sim$6000 random samples from $\mathcal{D}$. Furthermore, we introduce the prompt $h_C$ (see Table \ref{tab:prompt}), which serves as the model instructional component concatenated with $x_D$. The fine-tuning step is defined as 

\begin{equation}
    f(h_C, x_D) = (\Phi \circ f_0)(h_C, x_D, y),
\end{equation}

whereby $f_0$ represents the base model, $\Phi$ denotes the fine-tuning operation, while $f$ being our phase transition prediction model. Thus, phase transition predictions are inferred by

\begin{equation}
    y_p = f(h_C, x_D).
\end{equation}



\section{Experimental Results}

We first evaluate the performance of our two models on the PT dataset and compare it to the performance of two off-the-shelf models: Longformer and Llama 2.

\subsection{Statistics}
\begin{figure}[!ht]
\begin{center}
\includegraphics[width=2.5in]{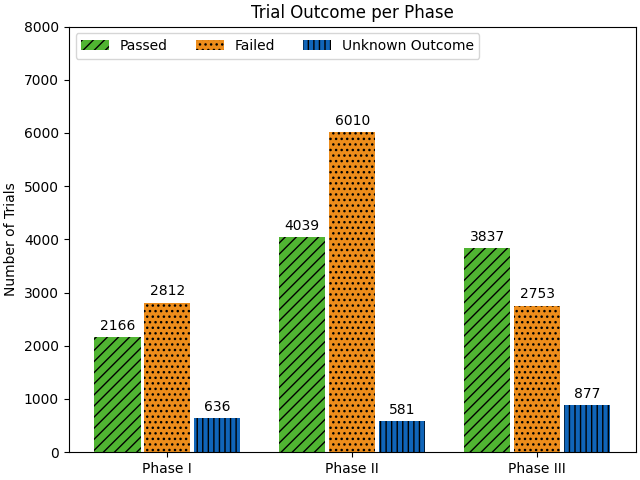}
\end{center}
\caption{Distribution of Passed, Failed, and Unknown Outcome Trials in our dataset. Phase II has the highest attrition rate while also having the most entered trials. Our reported outcome distribution differs from the classical literature~\citep{dimasi2010trends, hay2014clinical, kola2004can} as we use a novel labeling method for trial success. Still, it is in accordance with previous work using a similar approach as us \citep{feijoo2020key}.\label{fig:trial_outcome_per_phase}}
\end{figure}

\begin{figure}[!ht]
    \centering
    \subfigure[Distribution of the five different drug classes on the phases. The most prevalent type of drug in our dataset are New-Molecular Entities (NME), of which most enter the regulatory process in Phase II.]{
        \includegraphics[width=2.7in]{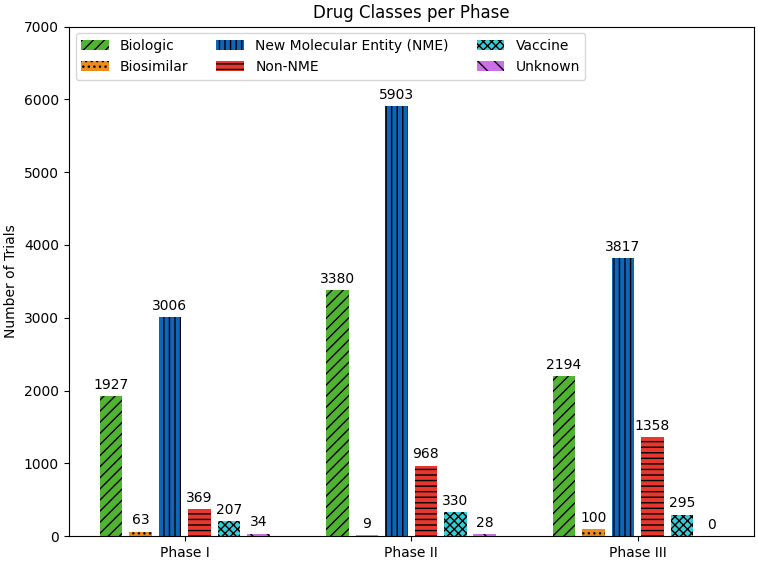}
    }
    \hspace{0.05\textwidth} 
    \subfigure[Pass-Fail Ratio of Trials by Drug Class. We report the highest attrition rates for drugs classified as unknown or unreported types, as well as for biologics and new molecular entities (NMEs). Conversely, biosimilars and non-NMEs exhibit high success rates, which aligns with expectations since they either already have market approval or are similar to drugs that do.]{
        \includegraphics[width=2.7in]{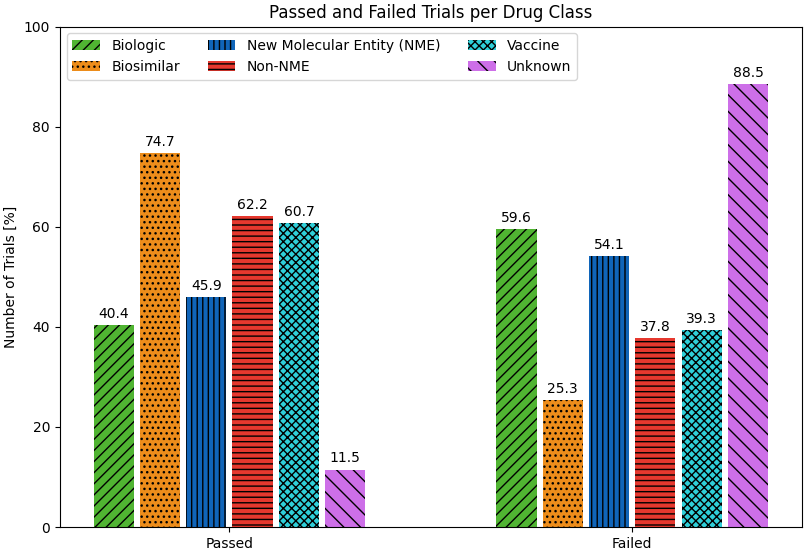}
    }
    \caption{Overview of Drug Classes and Their Impact on Clinical Trial Outcomes}
    \label{fig:side_by_side}
\end{figure}

Our dataset contains 21,617 labeled clinical trials and 2,094 unlabeled trials. Figure~\ref{fig:trial_outcome_per_phase} illustrates the distribution of trials across various phases, revealing that the majority of trials start in Phase II. The transition from Phase II to III experiences the highest attrition rate. These findings are consistent with the literature~\citep{feijoo2020key, hay2014clinical, kola2004can}. Conversely, trials in Phase III exhibit a higher likelihood of success. Figure~\ref{fig:side_by_side}A further analyzes the data distribution according to drug classes (for a detailed explanation, see \ref{sec:drug_classes}) over phases. Most tested drugs are New Molecular Entities (NMEs) and Biologics (blue and green bars), both of which have higher failure rates (see Figure~\ref{fig:side_by_side}B). Interestingly, trials with an unreported drug (purple bar) fail in $\sim$ 90\% of cases.

\subsection{Data Split}
\label{sec:data_split}
The following experiments are conducted on our PT dataset, introduced in Section \ref{sec:dataset-creation} and report a single run. We balanced the dataset to include an equal amount of successful and failed trials. The used dataset included 20,000 trials, which we split into 65\% training, 15\% validation, and 20\% test data. To fine-tune GPT-3.5 Turbo (see Section \ref{sec:CTP-LLM}), we have used a balanced subset of 6250 entries with the same split ratio. To avoid biasing the models, we sorted all trial entries chronologically by their last modification date. The training and validation sets include trials from 2005 to July 2022, while the test sets consist of trials dated from August 2022 onwards.

\begin{table}[!htp]
\begin{center}
\begin{tabular}{lll}
\toprule
\multicolumn{1}{c}{\bf Model}  &\multicolumn{1}{c}{\bf Accuracy}  &\multicolumn{1}{c}{\bf F1 Score} \\
\midrule
Longformer            & \multicolumn{1}{c}{0.515} & \multicolumn{1}{c}{0.508} \\
Clinical Longformer            & \multicolumn{1}{c}{0.644} & \multicolumn{1}{c}{0.599} \\
LLaMA 2 7B (AlpaCare)           & \multicolumn{1}{c}{0.518} & \multicolumn{1}{c}{0.515} \\
CTP-LLM          & \multicolumn{1}{c}{\textbf{0.667}} & \multicolumn{1}{c}{\textbf{0.665}} \\
BERT + RF         & \multicolumn{1}{c}{0.626} & \multicolumn{1}{c}{0.617} \\

\bottomrule
\end{tabular}
\end{center}
\caption{Comparison of model performances trained on trials from all phases.\label{tab:model-performance}}
\end{table}

\begin{table*}[!htbp]
\begin{center}
\begin{tabular}{llllllll}
\hline
\multicolumn{1}{c}{} & \multicolumn{2}{c}{\bf BERT + RF} & \multicolumn{2}{c}{\bf Clinical Longformer} & \multicolumn{2}{c}{\bf CTP-LLM}\\
\cmidrule(lr){2-3} \cmidrule(lr){4-5} \cmidrule(lr){6-7}
\bf Phase Transition & \bf Accuracy & \bf F1 Score & \bf Accuracy & \bf F1 Score & \bf Accuracy & \bf F1 Score \\
\midrule
Phase I to II  & 0.642 & 0.580 & 0.586 & 0.386 & \textbf{0.732} & \textbf{0.730} \\
Phase II to III & 0.580 & 0.533 & \textbf{0.610} & 0.501 &  0.604 & \textbf{0.600} \\
Phase III to Approval & 0.665 & \textbf{0.700} & 0.638 & 0.681 & \textbf{0.707} & 0.695 \\
\bottomrule
\end{tabular}
\end{center}
\caption{Here, we compare how well the models predict individual phase transitions when trained on data from all three phases simultaneously. We claim that CTP-LLM outperforms the other models in predicting Phase III to Approval because it effectively connects information from earlier phases. See Section \ref{sec:ablation}.}
\label{tab:per-phase-model-performance}
\end{table*}

\subsection{BERT+RF}
As mentioned in Section \ref{sec:synthesis}, we concatenated 11 high-quality attributes of the trial protocols, resulting in an 8,488-dimensional trial description feature vector on which we train an RF classifier (see Section \ref{sec:bert-rf}). This hybrid method enhances predictive accuracy by aggregating predictions from multiple decision trees. In our experiments, the BERT+RF model exhibited promising results, achieving an F1 score of 0.617 in predicting phase transitions across various clinical trials (see Table~\ref{tab:model-performance}). Even though its performance is similar to the Clinical Longformer, our approach can be trained nearly seven times faster and more memory efficient. It further excels in individual phase outcome prediction, slightly outperforming even CTP-LLM by 0.005 points in F1 score (see Table~\ref{tab:per-phase-model-performance}). The encoding of all 20.000 trial texts and training of the RF model takes approximately 20 minutes on a single RTX 4090 GPU.

\subsection{CTP-LLM} \label{sec:CTP-LLM}
We determined that a balanced set of $\sim$ 6000 samples (around 18M tokens) is sufficient to achieve a significant improvement in performance over the baselines while maintaining low costs. The fine-tuned model, CTP-LLM, outperforms BERT+RF by 0.048 on the F1 score if trained on trials from all phases simultaneously (see Table~\ref{tab:model-performance}) and by 0.061 when fine-tuned for Phase III trials (see Table~\ref{tab:model-III-performance}). The training took around 150 minutes through the dedicated API with associated costs of about \$100.

\subsection{Longformer}

We trained both the classical Longformer \citep{beltagy2020longformer} ($\sim$149M parameters) and a clinical version \citep{li2022comparative} ($\sim$207M parameters) for seven epochs, with a training duration of 150-180 minutes on a single RTX 4090 GPU. While the Clinical Longformer performs comparably to BERT+RF when trained on a single phase (see Table~\ref{tab:model-III-performance}), it is significantly outperformed by our models when trained across multiple phases (see Table~\ref{tab:per-phase-model-performance}).

\subsection{Llama 2 (AlpaCare)}
As the classical Llama 2 \citep{touvron2023llama} lacks medical understanding, we conducted our experiments using AlpaCare 7B, a pre-fine-tuned version of LLaMA 2 that was self-instructed on medical queries \citep{zhang2023alpacare}. Despite its considerably larger size than the Longformer architectures, its performance is inferior to all other evaluated models (see Tables~\ref{tab:model-performance} and \ref{tab:model-III-performance}). Even though better results can be expected when using the 13B and 70B versions, we decided only to utilize the 7B model due to resource limitations. 


\subsection{Ablation Study} \label{sec:ablation}
Previous approaches in CTOP \citep{feijoo2020key, fu2022hint, stallard2005decision, qi2019predicting} developed three separate models to predict the outcomes of Phase I, II, and III trials. In contrast, we train our models on trials from all three phases simultaneously. In reality, the same drug can be tested in multiple trials across all three phases at the same time, with the results of these trials influencing each other~\citep{feijoo2020key}. Therefore, we hypothesize that CTOP models should be trained across phases. To avoid look-ahead bias, we split the training and testing data according to protocol modification dates (see Section~\ref{sec:data_split}). Table \ref{tab:per-phase-model-performance} shows how well the models predict the three individual phase transitions when trained on data from all three phases. 

\paragraph{Single phase training versus all phases} 
To test our theory, we trained all models on $\sim$6000 Phase III trial samples to investigate how well they predict the Phase III to Approval transitions (see Table \ref{tab:model-III-performance}). To reduce cost, CTP-LLM was trained on only 2300 Phase III samples. Our BERT+RF model shows an improved performance of 11.67\% on the F1 score compared to the version trained on all three phases. 


\begin{table}[htbp]
\begin{center}
\begin{tabular}{lll}
\toprule
\multicolumn{1}{c}{\bf Model}  &\multicolumn{1}{c}{\bf Accuracy}  &\multicolumn{1}{c}{\bf F1 Score} \\
\midrule
Longformer            & \multicolumn{1}{c}{0.686} & \multicolumn{1}{c}{0.691} \\
Clinical Longformer            & \multicolumn{1}{c}{0.682} & \multicolumn{1}{c}{0.689} \\
LLaMA 2 7B (AlpaCare)           & \multicolumn{1}{c}{0.624} & 
\multicolumn{1}{c}{0.625} \\
BERT + RF         & \multicolumn{1}{c}{0.677} & \multicolumn{1}{c}{0.689} \\
CTP-LLM          & \multicolumn{1}{c}{\textbf{0.751}} & \multicolumn{1}{c}{\textbf{0.750}} \\

\bottomrule
\end{tabular}
\end{center}
\caption{Comparing the performance of our models (BERT+RF and CTP-LLM) to the baseline models when trained only on Phase III trials.\label{tab:model-III-performance}}
\end{table}
\paragraph{Influence of previous phases} The performance of CTP-LLM in predicting Phase~III to Approval is improved by 12.78\% in F1 score when only trained on Phase~III trial data (see Table \ref{tab:model-III-performance}), compared to CTP-LLM trained on data from all three phases (see Table \ref{tab:per-phase-model-performance}). This suggests a detrimental effect of cross-phase training on performance.  However, these results are not directly comparable, as the cross-phase training set contains only $\sim$1120 Phase~III trials, 49\% less than the dedicated Phase~III training set. 
If we train CTP-LLM on only these exact 1120 Phase III trials, the F1 score for predicting Phase III to Approval drops by 5.33\%. We conclude that the remaining \textbf{Phase I and Phase II trial data in CTP-LLM's training set holds valuable information on predicting the transition from Phase III to Approval.} The reverse scenario is also valid: trials from later regulatory phases (chronologically earlier) can influence outcome predictions for earlier phases (chronologically later). However, this bias is favorable. A drug can re-enter the regulatory process multiple times over several years or even decades~\citep{branch2014new, drews1997drug}. Consequently, understanding the performance of similar trials from later phases in the past is advantageous. 

These results indicate that an outcome prediction model based on LLMs benefits from training across multiple phases rather than being constrained to information from a single phase, whereas classification approaches based on shallow learning classifiers show enhanced performance when tailored to individual phases. 
\begin{figure}[!ht]
\begin{center}
\includegraphics[width=2.7in]{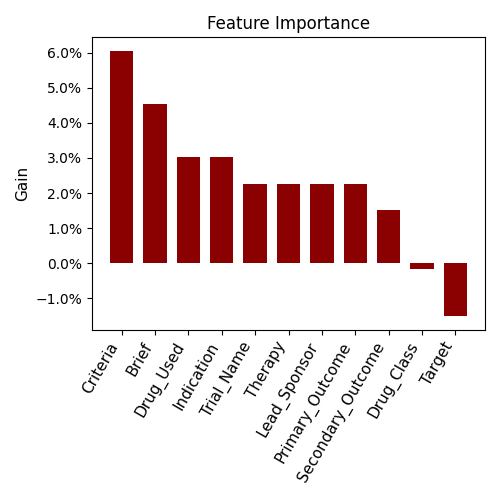}
\caption{ Drop-feature analysis reveals the relative importance of the eleven protocol features in predicting clinical trial outcomes using the CTP-LLM model. The participant selection criteria ("Criteria") stand out as the most critical factor, followed by the study description ("Brief") and the specific drug used. Notably, drug class exhibits a negative influence on the model's predictive power. See \ref{sec:feature_description} for definition of features. \label{fig:feature_importance}}
\end{center}
\end{figure}

\paragraph{Feature analysis}
Figure~\ref{fig:feature_importance} presents the findings of our drop-feature importance analysis. It reveals that participant selection criteria (Criteria) and study description (Brief) are the most influential factors in determining trial outcomes, with the drug itself ranking third. These findings highlight that study design variables significantly influence trial outcomes and cannot be neglected.

\section{Conclusion}
We have introduced a novel approach to predict clinical trial outcomes based on trial protocol descriptions automatically. To this end, we created a new labeled PhaseTransition dataset that includes protocol details for 20,000 trials, with labels derived from trials transitioning from one phase to the next. Using this dataset, we explored various transformer-based NLP methods to determine which approach most accurately predicts phase transitions. Our experiments show that a fine-tuned GPT-3.5 Turbo model (CTP-LLM) performs best when trained on trials from all three phases simultaneously;  effectively mirroring the real-world regulatory process by connecting information across phases. Additionally, our analysis reveals that participant selection criteria and study descriptions are the most critical factors in the model's decision-making process, outweighing drug characteristics.

\section*{Limitations}
Predicting the outcome of a clinical trial is an important topic. Nearly all previous work has focused on using ClinicalTrials.gov, the largest public database in the clinical trial domain. Even though ClinicalTrials.gov has more than 450,000 entries, we can only use a fraction of them, as we are limited by the quality of the data. Sponsors and investigators are responsible for submitting their data to ClinicalTrials.gov. In theory, the key information is registered at the start of the trial and updated regularly during its run, with the entry being closed as the study ends \citep{tse2018avoid}. In practice, the adherence to these requirements is inadequate, meaning that a large portion of entries lacks accurate and complete information \citep{tse2018avoid}. ClinicalTrials.gov stands as the largest database in its field, making it impractical or often impossible to verify data quality through cross-referencing with other sources. Furthermore, comprehending a clinical trial protocol demands profound expertise in the respective medical field. Given that the database contains trials across all medical domains, finding human experts with holistic medical knowledge to systematically and accurately assess data quality is exceedingly challenging. We are further limited by the second database, the BioMedTracker, as we can only use trials also featured in its much smaller collection. 
\par

Our work focuses on the binary classification of the trial phase transition outcome, an essential first step in a complex task. To further explain why a trial failure to transition is challenging. Our PT dataset maps drugs to trials for conditions and endpoints and then tracks how that drug transitions through phases. Thus, when looking at the data from an individual trial for an individual drug, it is not always clear if that particular drug transitioned to the next phase, unless we look at other trials that are available in the database. Second, while each trial has a protocol and results associated with it, the drugs often do not. Individual trials might be terminated for a multitude of reasons, including the success of a phase transition for the drug based on a number of other trials. Thus, looking at information from a single trial can rarely give all the information about why a drug might or might not transition into the next phase. We explored fine-tuning LLM to output reasons why a particular trial has failed (see Table~\ref{tab:reasoning} and Section~\ref{sec:reasoning}). 



\section*{Ethics Statement}
Our study provides a data-driven predictive framework for clinical trial outcomes, which will be a valuable tool for biomedical researchers, clinicians, and policy makers; however, it does not offer prescriptive advice on how to interpret or act upon these predictions. 
In other words, even if our model predict a trial to process from Phase to Phase, it shouldn't replace the clinical trial itself. Our model provides needed consultation with the goal to improve trial design and discover diagnostic factors related to trial success or failures. 
We also acknowledge the possibility of false positives or negatives, which may lead to erroneous conclusions. Our models should be used as a complementary tool alongside clinical judgment and not as a sole determinant in decision-making processes. We also realize  unexpected factors such as a pandemic might accelerate a drug's clinical trial progression. Our model is only a starting point to consider these more special situations. \par
Although our dataset relies on proprietary data obtained from BioMedTracker, it is crucial to note that all information published in our study is sourced from publicly available data. Only the labeling process utilizes restricted information, but we ensured no confidential data was disclosed in its raw form. Rather, this restricted data is utilized solely for the purpose of inference and labeling within our study. We emphasize our commitment to maintaining the confidentiality and integrity of proprietary information provided by BioMedTracker, while also adhering to ethical standards of transparency and accountability in our research methodology. 

\bibliography{custom}

\appendix

\section{Appendix}
\label{sec:appendix}

\begin{table}[hbp]
\begin{center}
\begin{tabular}{ll}
\toprule
\multicolumn{1}{l}{\bf Attribute}  &\multicolumn{1}{c}{\bf Symbol} \\
\midrule
Trial Name & \multicolumn{1}{c}{$x_N$} \\
Trial Brief & \multicolumn{1}{c}{$x_B$} \\
Drug Used & \multicolumn{1}{c}{$x_{DU}$} \\
Drug Class & \multicolumn{1}{c}{$x_{DC}$} \\
Indication & \multicolumn{1}{c}{$x_I$} \\
Target & \multicolumn{1}{c}{$x_T$} \\
Therapy & \multicolumn{1}{c}{$x_{Th}$} \\
Lead Sponsor & \multicolumn{1}{c}{$x_{S}$} \\
Recruitment Criteria & \multicolumn{1}{c}{$x_C$} \\
Primary Outcome & \multicolumn{1}{c}{$x_{PO}$} \\
Secondary Outcome & \multicolumn{1}{c}{$x_{SO}$} \\

\bottomrule
\end{tabular}
\end{center}
\caption{Attributes of the trial protocols used to create the trial description $x_D$. \label{tab:attributes}}
\end{table}

\subsection{Trial Protocol}
\label{sec:app_protocol}
A trial protocol is a comprehensive document that outlines the plan for conducting the trial. It details the study's objectives, design, methodology, statistical considerations, and organization. The protocol ensures participant safety and data integrity \citep{evans2010fundamentals, friedman2015fundamentals}. It is developed during the planning phase of the trial, typically by the trial's sponsor, who could be a pharmaceutical company,  a biotechnology firm, a medical device company, or an academic institution \citep{friedman2015fundamentals}. The key components of a protocol are the following:

\begin{itemize}
    \item \textbf{Objective}: Clearly defining the purpose of the trial, such as testing the efficacy and safety of a new drug or treatment ~\citep{evans2010fundamentals, friedman2015fundamentals}.
    \item \textbf{Design}: Describing the type of study (e.g., randomized, double-blind, placebo-controlled), including the number of participants, duration, and phases ~\citep{friedman2015fundamentals, getz2017trial}.
    \item \textbf{Eligibility Criteria}: Definition of the inclusion and exclusion criteria for participant selection to ensure a specific and appropriate study population ~\citep{friedman2015fundamentals, getz2017trial}.
    \item \textbf{Treatment Plan}: Specifying the interventions being tested, including dosages, administration methods, and schedules ~\citep{friedman2015fundamentals}.
    \item \textbf{Endpoints}: Lists the primary and secondary outcomes that will be measured to determine the intervention's effectiveness and safety ~\citep{friedman2015fundamentals, pazdur2008endpoints}.
    \item \textbf{Assessments and Procedures}: Outline of the tests, procedures, and data collection methods used throughout the study ~\citep{friedman2015fundamentals, getz2017trial}.
    \item \textbf{Statistical Analysis}: Details the statistical methods used to analyze the data and determine the significance of the results ~\citep{friedman2015fundamentals}.
    \item \textbf{Ethical Considerations}: Includes information on informed consent, participant rights, and measures to ensure participant safety and confidentiality ~\citep{friedman2015fundamentals, nardini2014ethics}.
    \item \textbf{Regulatory Compliance}: Ensures the trial adheres to relevant regulations and guidelines, such as those set by the FDA or EMA ~\citep{evans2010fundamentals, friedman2015fundamentals}.
\end{itemize}

According to ~\citet{friedman2015fundamentals}, a trial protocol typically has the following structure:

\begin{enumerate}[A.] 
    \item Background of the study
    \item Objectives
    \begin{enumerate}[1.] 
        \item Primary question and response variable
        \item Secondary questions and response variables
        \item Subgroup hypotheses
        \item Adverse effects
    \end{enumerate}
    \item Design of the study
    \begin{enumerate}[1.] 
        \item Study population
        \begin{enumerate}[(a)]
            \item Inclusion criteria 
            \item Exclusion criteria
        \end{enumerate}
        \item Sample size assumptions and estimates
        \item Enrollment of participants
        \begin{enumerate}[(a)]
            \item Informed consent
            \item Assessment of eligibility
            \item Baseline examination
            \item Intervention allocation (e.g., randomization method)
        \end{enumerate}
        \item Intervention(s)
        \begin{enumerate}[(a)]
            \item Description and schedule
            \item Measures of compliance
        \end{enumerate}
        \item Follow-up visit description and schedule
        \item Ascertainment of response variables
        \begin{enumerate}[(a)]
            \item Training 
            \item Data collection
            \item Quality control
        \end{enumerate}
        \item Assessment of Adverse Events
        \begin{enumerate}[(a)]
            \item Type and frequency 
            \item Instruments
            \item  Reporting
        \end{enumerate}
        \item Data analysis
        \begin{enumerate}[(a)]
            \item Interim monitoring, including data monitoring committee role
            \item  Final analysis
        \end{enumerate}
        \item Termination policy
    \end{enumerate}
    \item Organization
    \begin{enumerate}[1.]
        \item Participating investigators
        \begin{enumerate}[(a)]
            \item Statistical unit or data coordinating center
            \item Laboratories and other special units
            \item Clinical center(s)
        \end{enumerate}
        \item Study administration
        \begin{enumerate}[(a)]
            \item Steering committees and subcommittees
            \item Monitoring committee
            \item Funding organization
        \end{enumerate}
    \end{enumerate}
    \item[] Appendices
    \begin{enumerate}[]
            \item Definitions of eligibility criteria
            \item Definitions of response variables
            \item Informed Consent Form
        \end{enumerate}
\end{enumerate}

A complete trial protocol can have more than 100 pages. Due to their length and potential confidentiality, ClinicalTrials.gov primarily only hosts summaries or excerpts of the protocol document ~\citep{gresham2020clinicaltrials}. Table \ref{tab:data-point} shows an example of such a reduced protocol. We only included the most critical information while excluding sensitive data (e.g., names and phone numbers). 

\setlength\intextsep{2mm}{
\begin{table*}[t]
    \centering
    \scalebox{0.8}{%
        \begin{tabular}{|p{1.25\textwidth}|}
            \hline
            \vspace{-0.5\baselineskip}
            \textbf{System Prompt}\\
            \hline
            \\
            You are a medical expert who specializes in analyzing clinical trials. Your role is to help the user predict whether a clinical trial will progress to the next phase. 
            \newline\newline
            Answer only with `Yes' if it progresses to the next phase or `No' if it doesn't.


        \end{tabular}
        }
    \caption{LLM system prompt used for fine-tuning and inference.}
    
    \label{tab:prompt}
\end{table*}
}

\setlength\intextsep{2mm}{
\begin{table*}[!ht]
    \centering
    \scalebox{0.8}{%
        \begin{tabular}{|p{1.25\textwidth}|}
            \hline
            \vspace{-0.5\baselineskip}
            \textbf{Trial Description}\\
            \hline
            \\
    TRIAL NAME: Phase II - X2202; \newline
    BRIEF: The purpose of this study was to determine if BVS857 is safe, tolerable and increases thigh muscle thickness in patients with spinal bulbar and muscular atrophy (SBMA).; \newline
    DRUG USED: BVS857; \newline
    DRUG CLASS: New Molecular Entity (NME); \newline
    INDICATION: Spinal Bulbar Muscular Atrophy (SBMA, Kennedy's Disease, X-linked spinal muscular atrophy type 1); \newline
    TARGET: IGF-1R (Insulin-like Growth Factor-1 Receptor); \newline
    THERAPY: Monotherapy; \newline
    LEAD SPONSOR: Novartis Pharmaceuticals; \newline
    CRITERIA: Key Inclusion Criteria: - Genetic diagnosis of SBMA with symptomatic muscle weakness - Able to complete 2-minute timed walk - Serum IGF-1 level less than or equal to 170 ng/mL Key Exclusion Criteria: - Medically treated diabetes mellitus or known history of hypoglycemia - History of Bell's palsy - Treatment with systemic steroids \textgreater 10 mg/day (or equivalent dose); androgens or androgen reducing agents; systemic beta agonists; or other muscle anabolic drugs within the previous 3 months - History of cancer, other than non-melanomatous skin cancer - Retinopathy - Papilledema Other protocol-defined inclusion/exclusion criteria may apply;\newline
    PRIMARY OUTCOME: Number of Patients With Adverse Events (AEs), Serious Adverse Events (SAEs) and Deaths as a Measure of Safety and Tolerability; \newline
    SECONDARY OUTCOME: Mean Change From Baseline in Score on the Adult Myopathy Assessment Tool (AMAT) in Part B, Cohort 5
    \\

            \hline
            \vspace{-0.5\baselineskip}
            \textbf{Phase Transition}\\
            \hline
            \\
            No
            
            \\
        \end{tabular}
    }
    \caption{Example dataset entry. The field `Trial Description' is the input the model receives. It holds the most important information on the clinical trial while not exceeding the 4096 token limit. The field `Phase Transition' holds the associated label. In this example, the trial `X2202' did not transition from Phase II to Phase III. \label{tab:data-point}}
\end{table*}
}

\setlength\intextsep{2mm}{
\begin{table*}[!ht]
    \centering
    \scalebox{0.8}{%
        \begin{tabular}{|p{1.25\textwidth}|}
            \hline
            \vspace{-0.5\baselineskip}
            \textbf{Reasoning Examples}\\
            \hline
            \\
            Label: No. VTL-308 failed to meet its primary endpoint, all follow up studies were terminated.
            Predicted label: No. This clinical trial did not transition to the next phase. The reason is that the sponsor, Vital Therapies, Inc., discontinued the ELAD program, including the VTL-308 trial, due to lack of efficacy in previous studies.

            Label: No. Strategic
            \\
            Predicted label: No. This trial will not transition to the next phase. The reason is that the lead sponsor, Gilead Sciences, has not shown any involvement in oncology-related clinical trials involving CYT997 or glioblastoma multiforme.


            \\
            \hline
        \end{tabular}
    }
    \caption{Reasoning examples.  \label{tab:reasoning}}
\end{table*}
}

\subsection{Feature Description}
\label{sec:feature_description}

In this section, we provide detailed descriptions of the key features used in our clinical trial phase transition prediction model. These attributes are sourced from ClinicalTrials.gov and Biomedtracker, distilled into a streamlined version of a comprehensive clinical trial protocol, which then serve as the features on which the models base their prediction.

\begin{itemize}
    \item \textbf{Trial\_Name.} The official title of the clinical trial as listed on ClinicalTrials.gov. This often includes key information about the study purpose, population, phase, and intervention. 
    \item \textbf{Brief.} Refers to a shortened clinical trial description provided on ClinicalTrials.gov. It gives an overview of the study's objectives, design, and key characteristics.
    \item \textbf{Indication.} Describes the broad medical field or condition targeted by the trial, such as Solid Tumors, Breast Cancer, etc.
    \item \textbf{Target.} Specifies the medical condition or disease that the clinical trial is targeting.
    \item \textbf{Drug\_Used.} Refers to the specific drug or drugs being tested in the clinical trial. 
    \item \textbf{Drug\_Class.} See \ref{sec:drug_classes}.
    \item \textbf{Therapy.} Describes the type of therapeutic approach being employed in the trial. Examples are Monotherapy, Targeted Therapy, etc.
    \item \textbf{Lead\_Sponsor.} The organization or individual responsible for conducting the clinical trial and ensuring it adheres to regulatory requirements.
    \item \textbf{Criteria.} The inclusion and exclusion criteria determine which participants are eligible to join the clinical trial.
    \item \textbf{Primary\_Outcome.} The main result that the trial is designed to measure is usually specified as an endpoint in the study protocol. Primary outcomes are critical for assessing the efficacy and safety of the intervention.
    \item \textbf{Secondary\_Outcome.} Additional results measured in the trial to provide more information on the intervention's effects. Secondary outcomes can offer insights into other benefits or risks associated with the treatment and help support the primary outcome findings.
\end{itemize}

\subsection{Overview of Drug Classes}
\label{sec:drug_classes}
Within our dataset, we find clinical trials focusing on drugs categorized into five groups: Biologic, Biosimilar, New Molecular Entity, Non-New Molecular Entity, and Vaccine. Additionally, some trials are categorized as "Unknown." In the subsequent sections, we provide comprehensive explanations for each of these classifications:\\

\textbf{Biologics} or biopharmaceuticals are medicinal products derived from living organisms or their cells. These drugs typically comprise proteins, sugars, nucleic acids, or combinations of these substances \citep{dranitsaris2011biosimilars, morrow2004defining}. In contrast to traditional pharmaceuticals, their production requires more sophisticated methods, such as recombinant DNA technology, controlled gene expression, and cell culture techniques. Their structures are substantially more complex and larger in molecular size compared to traditional small-molecule drugs \citep{dranitsaris2011biosimilars, geynisman2017biosimilar, morrow2004defining}. Furthermore, they are designed to target specific components of the body's biological pathways, leading to fewer side effects and effectiveness in treating diseases that are difficult to manage with conventional drugs. However, they are often sensitive to temperature and storage conditions, and due to their biological nature, they more easily trigger immune responses \citep{dranitsaris2011biosimilars, morrow2004defining}.
    
\textbf{Biosimilars} are a type of biological product that are highly similar to an already approved biological drug, even though they are generally not interchangeable \citep{dranitsaris2011biosimilars, geynisman2017biosimilar}. Driven by market demands, competitors are trying to develop their own version of innovative drugs, but without access to manufacturing data from the brand company, creating an identical product is virtually impossible \citep{dranitsaris2011biosimilars}. Therefore, the industry has focused on developing biological agents that are clinically and biologically comparable to their target product. To receive regulatory approval, the company has to demonstrate pharmacokinetic comparability to the original
product, whereby the FDA assesses the degree of comparability on a case-by-case basis \citep{dranitsaris2011biosimilars, geynisman2017biosimilar}.

\textbf{New Molecular Entity (NME)} is a novel drug whose active ingredient is a chemical substance that is marketed for the first time in the United States~\citep{branch2014new, eaglstein2011new}. Per definition, NMEs refer to both chemical drugs and biologics~\citep{branch2014new}. These innovative compounds represent novel therapeutic options designed to target diseases or conditions that current treatments cannot adequately address. The approval of NMEs by the FDA marks a significant milestone, as it indicates the introduction of a potentially groundbreaking treatment into the market~\citep{schuster2005drugs, feijoo2020key}. These drugs can significantly impact public health by offering new solutions for previously unmet medical needs. However, due to their novelty, they have a higher failure rate compared to other drugs~\citep{feijoo2020key}.

\textbf{Non-NME} or Non-New Molecular Entities refer to drugs that do not introduce a novel active ingredient but rather involve modifications or new applications of already approved substances~\citep{branch2014new}. These can include new formulations, such as extended-release versions, new combinations of existing drugs, or new therapeutic indications for previously approved drugs~\citep{branch2014new, drews1997drug}. While they utilize existing active ingredients, Non-NMEs offer significant benefits by improving patient compliance, expanding therapeutic options, and potentially reducing development costs and time~\citep{drews1997drug}.

\textbf{Vaccines} are biological preparations designed to stimulate the body's immune system to recognize and fight against specific pathogens, such as viruses or bacteria, thereby preventing or reducing the severity of infectious diseases~\citep{gebre2021novel, han2015clinical}. They can be composed of chemical substances or biological components, such as proteins, polysaccharides, or genetic material~\citep{gebre2021novel, han2015clinical}. In the context of regulatory approval, a vaccine's efficacy is measured by its ability to produce a strong and lasting immune response and its effectiveness in preventing the targeted disease~\citep{han2015clinical, jager2002clinical}.

The \textbf{Unknown} drug type refers to trials where the specific type or classification of the drug being studied is not disclosed or not provided in the trial registration information. This may occur for various reasons, such as incomplete or missing data in the trial record or because the drug being studied has not been categorized or classified into a specific drug type~\citep{stergiopoulos2019evaluating, tse2018avoid}.


\subsection{Reasoning Experiment} \label{sec:reasoning}
Leveraging an LLM for predicting trial phase transitions allows us to converse with the model about its decision-making process. Thereby, one could gain information on which factors influenced the prediction and identify flaws in trial protocols.\par
Several terminated trials on ClinicalTrial.gov state the reason for termination. We augmented our PT dataset by concatenating this statement with the `No'-label, while the `Yes'-label remains unchanged. To test the reasoning capabilities of CTP-LLM, we fine-tuned another version of the model on the modified `No'-label, instructing the model to give a reason for its decision every time it predicts that a trial will not perform a phase transition.\par
During evaluation, it became evident that the fine-tuned reasoning model was biased toward attributing trial termination to safety and efficacy concerns. However, there are also some cases where the model exhibits interesting reasoning abilities and indicates prior knowledge about a specific trial program or sponsor. Examples are given in Table~\ref{tab:reasoning}.



\end{document}